\documentclass[letterpaper, 10 pt, conference]{ieeeconf}

\IEEEoverridecommandlockouts
\usepackage[english]{babel}
\usepackage{amsmath} % assumes amsmath package installed
\usepackage{amssymb}  % assumes amsmath package installed
\usepackage{graphicx}
\usepackage{subfigure}
\usepackage{caption}
\usepackage{multirow}
\usepackage{amsfonts}
\usepackage{hyperref}
\usepackage{tabularx}
\usepackage{algorithm,algpseudocode}
\usepackage{bm}
\usepackage{breakurl}
\usepackage{enumerate}
\usepackage{verbatim}
\usepackage{float}
\usepackage{siunitx}
\usepackage{mdframed}
\usepackage{adjustbox}
\usepackage{cleveref}
\usepackage{authblk}
\usepackage{color}
\usepackage{here} 
\usepackage{soul}
\usepackage{tabularx}
\newcolumntype{b}{X}
\newcolumntype{s}{>{\hsize=.3\hsize}X}
\newcolumntype{x}{>{\hsize=.5\hsize}X}
\usepackage{tikz}
\usepackage{ dsfont }
\usetikzlibrary{shapes.geometric, arrows}

\usepackage{array}
\newcommand{\camf}{\mathcal{C}}
\newcommand{\robotf}{\mathcal{B}}
\newcommand{\worldf}{\mathcal{I}}

\newcommand{\vece}{\mathbf{e}}

\newcommand{\vecq}{\mathbf{q}}

\newcommand{\vecu}{\mathbf{u}}

\newcommand{\vecx}{\mathbf{x}}

\newcommand{\matU}{\mathbf{U}}
\newcommand{\matR}{\bm{\mathit{R}}}

\newcommand{\matQ}{\mathbf{Q}}

\newcommand{\massQ}{m_{Q}}

\newcommand{\angvel}{\mathbf{\Omega}}

\newcommand{\robotpos}[1]{\vecx_{#1}}
\newcommand{\robotrot}[1]{\matR_{#1}}

\newcommand{\robotvel}[1]{\dot{\vecx}_{#1}}
\newcommand{\robotacc}[1]{\ddot{\vecx}_{#1}}
\newcommand{\robotangvel}[1]{\angvel_{#1}}

\newcommand{\quat}{\vecq}

\newcommand{\realnum}[1]{\mathbb{R}^{#1}}
\newcommand{\SOthree}{SO(3)}

\newcommand{\norm}[1]{\left\lVert#1\right\rVert}

\newcommand{\twonorm}[1]{\left\lVert#1\right\rVert_2}
\newcommand{\prths}[1]{\left(#1\right)}

\usepackage{cite}

\usepackage{wasysym}  % symbols
\author{Guanrui Li$^*$, Alex Tunchez$^*$, and Giuseppe Loianno
\thanks{$^*$These authors contributed equally.}
\thanks{The authors are with the New York University, Tandon School of Engineering, Brooklyn, NY 11201, USA. {\tt\footnotesize email: \{lguanrui, atunchez, loiannog\}@nyu.edu}.}
\thanks{This work was supported by the NSF CAREER Award 2145277, the NSF CPS Grant CNS-2121391, the Technology Innovation Institute, Qualcomm Research, Nokia, and NYU Wireless.}
}

\title{\LARGE \bf Learning Model Predictive Control for Quadrotors}

\begin{document}

\maketitle
\thispagestyle{empty}
\pagestyle{empty}

\begin{abstract}
Aerial robots can enhance their safe and agile navigation in complex and cluttered environments by efficiently exploiting the information collected during a given task. In this paper, we address the learning model predictive control problem for quadrotors. We design a learning receding--horizon nonlinear control strategy directly formulated on the system nonlinear manifold configuration space $SO(3)\times\mathbb{R}^3$. The proposed approach exploits past successful task iterations to improve the system performance over time while respecting system dynamics and actuator constraints. We further relax its computational complexity making it compatible with real-time quadrotor control requirements. We show the effectiveness of the proposed approach in learning a minimum time control task, respecting dynamics, actuators, and environment constraints.
Several experiments in simulation and real-world setup validate the proposed approach.
\end{abstract}

\IEEEpeerreviewmaketitle
\section{Introduction}
Micro Aerial Vehicles (MAVs) such as quadrotors have become very popular platforms to help humans solve a wide range of time-sensitive problems in constrained outdoor and indoor environments including logistics, search and rescue for post-disaster response, and more recently during COVID-19 pandemic reconnaissance and monitoring.  These time-sensitive tasks  would require  robots  to  make  fast  decisions  and  agile  maneuvers in  uncertain, cluttered, and dynamic environments by intelligently exploiting the environment information to improve their performances over time. In this work, we investigate a Learning Model Predictive Control (LMPC) for quadrotors exploiting  past successful task iterations to improve  its  task performance over time while respecting system dynamics and actuator constraints. 

Several works have investigated the use of MPC for quadrotor control with perception and actuator constraints~\cite{Falanga2018,Bicego2020,Jacquet2021,LiTunICRA2021}. Other works~\cite{Bouffard2012, Torrente2021} improve MPC performance by refining the system dynamics in a data-driven fashion. Conversely, approaches such as~\cite{Williams2017} approximate the system dynamics directly using neural networks. However, these methods are quite computationally expensive since they rely on a sampling approach that is generally performed in parallel using Graphics Processing Units (GPUs).
Iterative learning control techniques~\cite{Bristow2006} have been successfully combined with MPC in a Batch Model Predictive Control (BMPC) approach to control chemical processes~\cite{Lee2007} and refine their performances over multiple task iterations. In~\cite{Rosolia2018,ROSOLIA20173142} a learning MPC approach is proposed and applied to ground vehicles. In this approach, the vehicle collects the states and their corresponding costs, across multiple successful iterations of the same task. The vehicle learns from the collected data to explore new ways to decrease cost in the same task as long as it maintains the ability to reach a state that has already been demonstrated to be safe during previous iterations. The approach does not require a reference trajectory as in previously mentioned works; thus, it is especially versatile and useful during tasks where the desired trajectory is not known or difficult to compute due to the system complexity or parameter uncertainty.  This is the typical case of drone racing competitions~\cite{Moon2019,Guerra2019,Foehn2020,MohtaJFR2018} that have recently inspired researchers to design autonomy algorithms with the goal to grant vehicles the ability to execute agile maneuvers with superior performances compared to human controlled vehicles.
Therefore, inspired by~\cite{Rosolia2018,ROSOLIA20173142}, we propose an LMPC approach for quadrotors. Common multi-rotor platforms including quadrotors evolve on the nonlinear manifold configuration space $SO(3)\times \realnum{3}$ making the LMPC problem substantially different and more complex for these types of systems compared to ground vehicles. We address the challenges of building a safety set that includes members of the rotation group $SO(3)$. Also, we consider an appropriate numerical integration approach for the group elements to ensure that the forward integration results adhere to the $SO(3)$ structure once employed in the discrete MPC formulation. In addition, we carefully add several design considerations to make the approach compatible with real-time control requirements of small aerial robots and show its feasibility in a learning minimum time control problem.
\begin{figure}[t]
    \centering
    \includegraphics[width=\columnwidth]{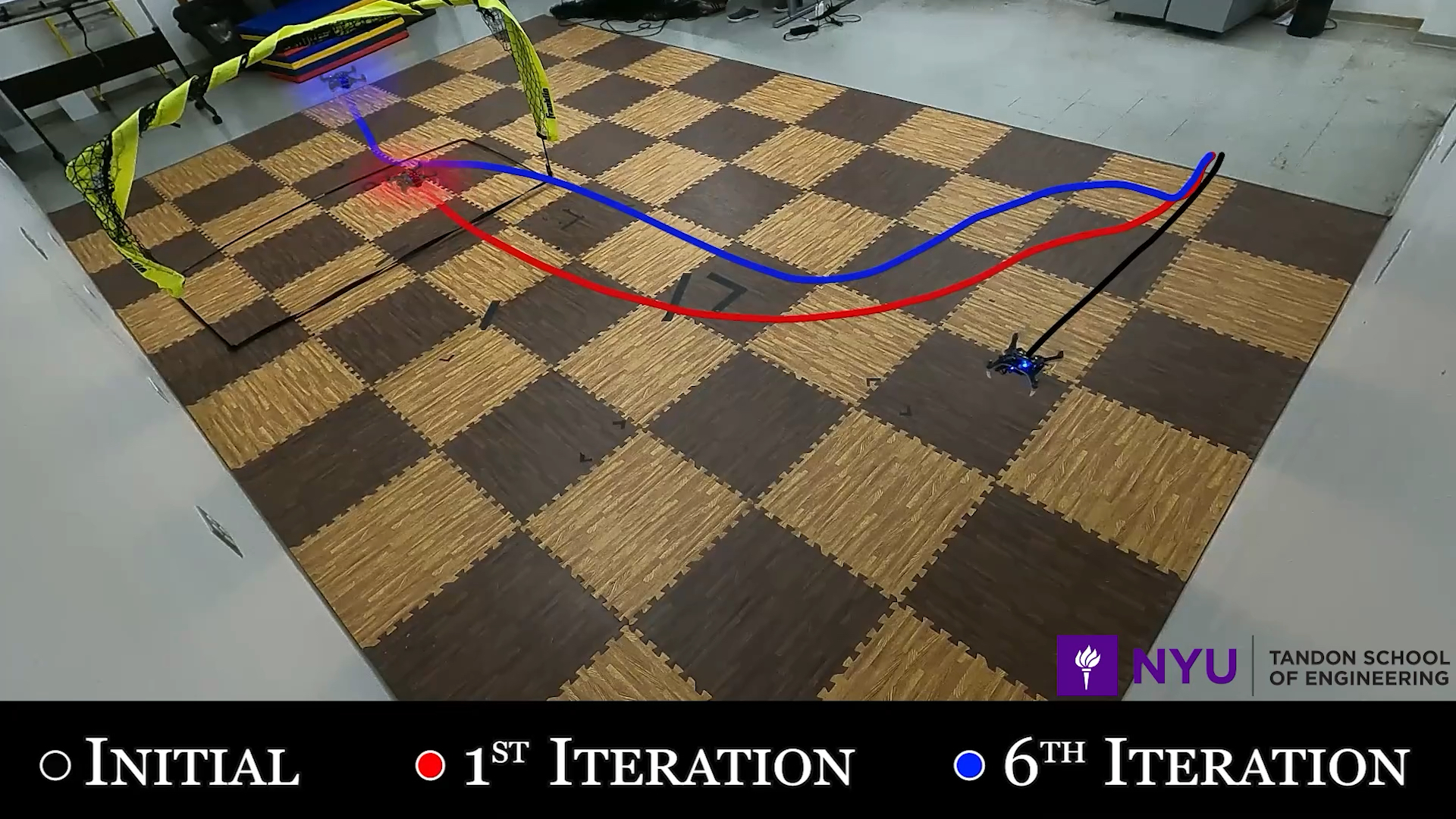}
    \caption{The LMPC iterates over the same task while learning  minimum time trajectories and improving its performances.}
    \label{fig:intro}
    \vspace{-10pt}
\end{figure}

The contribution of this paper is threefold. First, we propose an LMPC for quadrotors. We tackle the challenges related to transitioning this class of MPC to multi-rotor aerial systems. The data collected across multiple iterations incorporates elements of the nonlinear manifold $SO(3)$. Furthermore, this aspect also makes the numerical MPC integration substantially more complex since at each integration step it is necessary to guarantee that the obtained elements still lie in the rotation group. Second, we relax the computational  complexity of the proposed approach making it suitable for real-time quadrotor control applications. Finally, we show a specific instance of the proposed method in a learning minimum time control task. Several simulation and experimental results show the ability of our approach to successfully exploit collected data to improve system execution time performances.

The paper is organized as follows. In Section~\ref{sec:lmpc}, we review the MPC and LMPC approaches. In Section~\ref{sec:lmpcquadrotors}, we specifically address the challenges and show how to design LMPC for quadrotors, whereas, in Section~\ref{sec:optimaltime}, we consider a particular instance of this approach employed to solve a minimum time control task. Section~\ref{sec:experimental_results} presents our experimental results and Section~\ref{sec:conclusion} concludes the paper.

%can be employed in a variety of real-world applications such as inspection, maintenance, search and rescue, and package delivery. The versatility and mobility of UAVs has boosted the need to create autonomous aerial robots interacting with the environment, like transporting and manipulating objects. Aerial transportation can offer a faster and more versatile solution compared to ground transportation in congested urban environments and in post-disaster scenarios where aerial robots can deliver supplies. 

\section{Overview}\label{sec:lmpc}
In this section, we provide a brief review of the MPC formulation and how it transforms into LMPC.

\subsection{Model Predictive Control}~\label{sec:mpc}
MPC is a predictive control method that finds a sequence of system inputs, $\matU = \{\vecu_{0},\vecu_{1},\cdots,\vecu_{N-1}\}$ with $\vecu_k\in\realnum{n_u}$, within a fixed time horizon of $N$ steps. It optimizes a given objective function -- with a running and terminal cost $h(\cdot,\cdot) \text{ and } Q(\cdot)$ respectively -- while accounting for constraints and system dynamics
\begin{equation}
    \begin{split}
        \min_{\vecu_{0},\vecu_{1},\cdots,\vecu_{N-1}}& ~\sum_{k=0}^{N-1}  h\left(\vecx_{k},\vecu_{k}\right) + Q\left(\vecx_N\right)\label{eq:MPC_Form},\\ 
        \text{s.t.}&~\vecx_{k+1} = \mathbf{f}\left(\vecx_{k},\vecu_{k}\right),~\forall k = 0,\cdots,N-1\\
        &~\vecx{}_0 = \vecx{}(t_0),\\
&~\mathbf{g}\left(\vecx_{k},\vecu_{k}\right)\leq 0,
\end{split}
\end{equation}
where $\vecx_{k+1} = \mathbf{f}\prths{\vecx_k,\vecu_k}$ represents the system dynamics and $\vecx_k\in\realnum{n_x}$ is the system state. The optimization occurs with initial condition $\vecx_0$ while respecting system dynamics $\mathbf{f}(\vecx, \vecu)$ and additional state and input constraints $\mathbf{g}\left(\vecx,\vecu\right)$. %The formulation of eq.~(\ref{eq:MPC_Form}) is a nonlinear constrained optimization problem. The model dynamics are discretized with a time step $dt$ for the fixed time horizon $t_h$ into $\vecx_k,~k = 0,\cdots,N$ and $\vecu_k,~k= 0,\cdots,N-1$. This results in a fixed horizon length $\left[0,N\right]$, where, $N = \frac{t_h}{dt}$, and $N$ is selected based on performance for the real-time implementation.% We implement a receding horizon MPC strategy to finally define the LMPC:

\subsection{Learning Model Predictive Control}\label{sec:LMPC}
In this section, we review the LMPC, an iterative learning model predictive control method that can improve task performance over many trials~\cite{Rosolia2018}. The task is repeated at each iteration starting from the same initial state, $\mathbf{x}_s$. The task is considered complete when the system reaches a global terminal state, $\mathbf{x}_F$, without violating constraints. During each task iteration, the system records the state and cost and uses the recorded data to ensure control convergence to a locally optimal solution. We denote state and cost recorded during the $j^\text{th}$ iteration as the $j^\text{th}$ closed-loop trajectory. In the following, we first introduce preliminary concepts for LMPC such as safety set and its corresponding terminal cost function. Then we will introduce the LMPC formulation.

\subsubsection{Safety Set}
The states and inputs for a trajectory at each $j^\text{th}$ iteration is defined as
\begin{equation}
    \vecx{}^j = \left[\vecx^j_0,\vecx^j_1,...,\vecx^j_{T^j}\right],~
    \vecu{}^j = \left[\vecu^j_0,\vecu^j_1,...,\vecu^j_{T^{j}{-1}}\right],
\end{equation}
where $T^j$ is the time stamp when the task is completed at the $j^\text{th}$ iteration. The recorded data points from each iteration generate a sampled safety set
\begin{equation}
    SS^j = \left\{ \bigcup\limits_{i\in M^j}~\bigcup\limits_{t=0}^{T^i}x^i_t\right\} \label{SafetySet},
\end{equation}
where $M^j$ is a set of indexes that represents the iterations that successfully completed the task. In short, $SS^j$ is a set for the $j^\text{th}$ iteration which contains the recorded state trajectories from previous successfully completed tasks.
 
\subsubsection{Cost Function}
The cost-to-go for the state $\vecx_t^j$ at time $t$ of the $j^{th}$ iteration in the safety set is defined as
 \begin{equation}
     J^j_{t\xrightarrow{}T^j}(\vecx{}_t^j) = \sum_{k=t}^{T^j} \label{eq:cost2go} h\left(\vecx{}^j_{k},\vecu{}^j_{k}\right).
 \end{equation}
The cost-to-go for any state can be determined with eq.~(\ref{eq:cost2go}) which is needed to satisfy LMPC formulation as discussed in Section~\ref{sec:lmpc-formulation}. Furthermore, the optimal cost of a given state is obtained from the minimum cost-to-go across all previous successful iterations
\begin{align}
    Q^j(\vecx{}) = \begin{Bmatrix}
    \displaystyle\min_{(i,t)\in F^j(\vecx{})} J^i_{t\xrightarrow{}T^i}(\vecx{}),&\text{if} ~ \vecx{} \in SS^j \\
    +\infty, &\text{if} ~ \vecx{} \not\in SS^j
    \end{Bmatrix}, \label{eq:mincost2go}
\end{align}
\begin{align*}
    F^j(\vecx{}) = 
    \begin{Bmatrix}
        (i,t) | i \in [0,j], t \geq 0~\text{with} ~ \vecx{} = \vecx{}^i_t,
        ~\text{for}~ \vecx{}^i_t \in SS^j
    \end{Bmatrix}.
\end{align*}
Eq.~(\ref{eq:mincost2go}) assigns every state the respective minimum cost-to-go across all iterations.

\subsubsection{LMPC Formulation}\label{sec:lmpc-formulation}
The LMPC builds upon the general MPC formulation as described in Section~\ref{sec:mpc} by appending a safety set constraint for the terminal state and assigning eq.~\eqref{eq:mincost2go} as the terminal cost
\begin{subequations}
\label{eq:LMPCform}
    \begin{equation}
        \begin{split}
            J^{\scalebox{0.5}{$LMPC$},j}_{t\xrightarrow{}t+N}\left(\vecx_t^j\right)=\min_{\vecu_{t|t},...,\vecu_{t+N-1|t}}   \sum_{k=t}^{t+N-1} \hspace{-5pt}h&\left(\vecx_{k|t},\vecu_{k|t}\right)\\ &+~Q^{j-1}\left(\vecx_{t+N|t}\right)\label{eq:LMPCcost},\\ 
        \end{split}
    \end{equation}
    \begin{align}
        \text{s.t.}&
        ~\vecx_{k+1|t} = \mathbf{f}\left(\vecx_{k|t},\vecu_{k|t}\right),~\forall k \in [t,\cdots,t+N-1] \label{eq:LMPCdynamics},\\
        &\mathbf{g}\left(\vecx_{k|t},\vecu_{k|t}\right)\leq0,\label{eq:LMPC_lincon}\\
        &\vecx_{t+N|t}\in SS^{j-1}, \label{eq:SSconstraint}\\
        &\vecx_{t|t} = \vecx^j_t\label{eq:LMPC_initcond}.
    \end{align}
\end{subequations}
The optimal control problem in eq.~(\ref{eq:LMPCform}) computes a solution for the $j^\text{th}$ iteration at a given time stamp $t$ of the task, over a finite horizon $N$.
The eqs.~(\ref{eq:LMPCdynamics}),~(\ref{eq:LMPC_lincon}),~(\ref{eq:LMPC_initcond}) define the system dynamics, state, and input constraints, and initial condition of the system, respectively. The safety set constraint in eq.~\eqref{eq:SSconstraint} forces the terminal state $\vecx_{t+N|t}$ to visit a discrete safety set state in $SS^{j-1}$. In principle, this guarantees a closed-loop solution that pushes the system to a final state $\mathbf{x}_F$.
%every point in the safety set has a corresponding control policy which has proven to complete the desired task. Additionally, the cost-to-go for the corresponding discrete predicted state is included. During, real-time implementation a receding-horizon control strategy is used. Only the first optimal input in applied over $N$, and the problem is repeatedly solved for consecutive time stamps, $t$.

\section{LMPC for Quadrotors}~\label{sec:lmpcquadrotors}
In this section, we address the challenges related to designing LMPC for quadrotors. These systems evolve on complex nonlinear manifold configuration spaces $SO(3)\times \realnum{3}$ thus making the formulation and solution of the LMPC substantially more complex than ground vehicles studied in~\cite{Rosolia2018}. This induces the need for appropriate numerical integration techniques of the system dynamics preserving the structure of the configuration space over time. Furthermore,  the controller should run at an acceptable rate for real-time control of the quadrotor. The discrete property of the safety set, $SS^j$, makes the LMPC a Nonlinear Mixed Integer Programming (NMIP) problem due to eq.~(\ref{eq:SSconstraint}). This formulation is computationally expensive and incompatible with real-time applications. Hence in the following, we propose a convex approximation approach in Section~\ref{sec:convex-ss} in order to transfer the safety set constraints to a linear constraint and speed up the optimization.
In addition, the size of the safety set $SS^j$ is $n_x \times (\sum^j_{i=0} \tilde{f}T^i)$ and depends on the size of a state $n_x$, sample frequency $\tilde{f}$, and time $T^i$ of completion for successful $i^\text{th}$ iteration. Based on the formula, we see that the size of the safety set can grow very quickly across iterations. Therefore, it is beneficial to create a sparser safety set for computational reasons. We propose to select a subset of $SS^j$ as a local safety set to reduce the computational burden. Finally, we consider the limitation of the hardware platform. We incorporate in our formulation actuator constraints to guarantee the task feasibility as well as safety of the quadrotor platform. 

%First, we provide an overview on the quadrotor system dynamics. Subsequently, we introduce the safety set as shown in Section~\ref{sec:lmcp} and its local and convex approximations that contribute to reduce the computational burden to enable our approach to work in real-time. Finally we present the final formulation of LMPC on quadrotor system.

\subsection{System Dynamics}
\begin{figure}[!t]
  \centering
    \includegraphics[width=0.9\columnwidth]{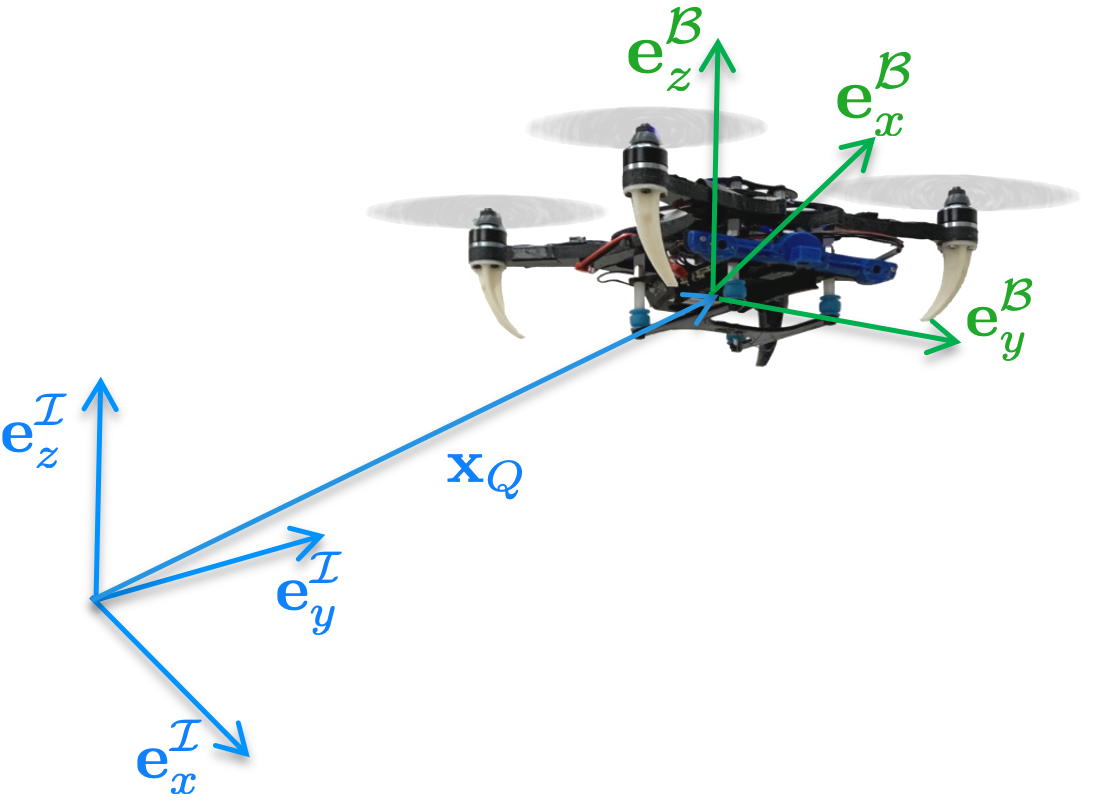}
  \caption{System  convention with $\mathcal{I}$ and $\mathcal{B}$ denoting inertial and body frames respectively.
  \label{fig:system_overview}}
  \vspace{-20pt}
\end{figure}

The system overview is presented in Fig.~\ref{fig:system_overview}. The relevant variables used in our paper are stated in Table~\ref{tab:notation}. We use two different coordinate frames with axes \{$\vece{}^{\mathcal{I}}_x,\vece{}^{\mathcal{I}}_y,\vece{}^{\mathcal{I}}_z$\} and \{$\vece{}^{\mathcal{B}}_x,\vece{}^{\mathcal{B}}_y,\vece{}^{\mathcal{B}}_z$\}, to denote the inertial frame and the quadrotor's body frame, respectively. The relevant variables for our model are defined in Table~\ref{tab:notation}. We consider $\vecx{} = [\robotpos{Q}^{\top}, \robotvel{Q}^{\top},\quat{}^{\top}]^\top$ and $\vecu = [f, \angvel{}^{\top}]^\top$ the quadrotor state and input vectors, respectively.  The state vector includes the quaternion $\quat{} = \left[q_w,q_x,q_y,q_z\right]^\top$ to describe the quadrotor's orientation in the $\mathcal{I}$ frame. We employ this orientation parameterization instead of a rotation matrix so that we can reduce our state space size and consequently speed up our LMPC as mentioned in Section~\ref{sec:lmpc}.

The system dynamics are
\begin{align}
\frac{d}{dt}\robotpos{Q} = \robotvel{Q}&,\hspace{1em}\frac{d}{dt}\robotvel{Q} = \frac{1}{\massQ}\left(f\robotrot{}\vece{}^{\mathcal{I}}_z - g\vece{}^{\mathcal{I}}_z\right),\label{eq:qd_vel}\\
\frac{d}{dt}\quat{} &= \frac{1}{2}\Lambda\left(\angvel{}\right)\cdot \quat{}, \label{eq:quatdot}
\end{align} 
where $\Lambda\left(\angvel{}\right)$ is a skew-symmetric matrix of the quadrotor angular velocity~\cite{Sol2017QuaternionKF} with $\angvel = \left[\Omega_x, \Omega_y,\Omega_z\right]^\top$.% defined as
%\begin{equation}
%   \Lambda(\angvel{}) = 
%   \begin{bmatrix}
%     0 & -\Omega_x & -\Omega_y & -\Omega_z \\
%     \Omega_x & 0 & \Omega_z & -\Omega_y \\
%     \Omega_y & -\Omega_z & 0 & \Omega_x \\
%     \Omega_z & \Omega_y & -\Omega_x & 0
%   \end{bmatrix}.
%\end{equation}
\begin{table}[!b]
\caption {Notation table.\label{tab:notation}} 
\centering
\begin{tabularx}{0.48\textwidth}{>{\hsize=0.58\hsize}X >{\hsize=1.42\hsize}X}
    \hline\hline
 $\worldf$, $\camf$, $\robotf$ & inertial, camera, robot frame\\
 $\massQ \in \realnum{}$ &  mass of robot\\
 $\robotpos{Q},\robotvel{Q}, \robotacc{Q}\in\realnum{3}$ &  position, velocity, acceleration of robot in $\worldf$\\
 $\robotrot{}\in\SOthree$&rotation matrix of robot with respect to $\worldf$ \\
  $\robotangvel{}\in\realnum{3}$& angular velocity of robot in $\robotf$\\
$f\in\realnum{}$&total thrust\\
 $g\in\realnum{}$& gravity constant\\
    \hline\hline
\end{tabularx}
\vspace{-1em}
\end{table}

The total thrust is the lift generated by each of the rotors $f=f_1+f_2+f_3+f_4$ along $\vece{}^{\mathcal{B}}_z$. Subsequently, it is necessary to apply a rotation induced by $\quat{}$ onto $\vece{}^{\mathcal{I}}_z$ by using
\begin{equation}
    \quat{}\odot\vece{}^{\mathcal{I}}_z = \robotrot{}\vece{}^{\mathcal{I}}_z. \label{eq:rotmat}
\end{equation}

\subsection{Discretization and Numerical Integration}\label{sec:numintegration}
The eqs.~\eqref{eq:qd_vel}-\eqref{eq:quatdot} represent the continuous system dynamics of a quadrotor which can be written as $\dot{\vecx} = \mathbf{f}(\vecx,\vecu)$. In order to incorporate them in the discrete time LMPC formulation, we apply $4^\text{th}$ order Runge-Kutta method to numerically integrate $\dot{\mathbf{x}}$ over the sampling time $dt$ given the state $\vecx_{k}$ and input $\vecu_{k}$ as
\begin{equation}
    \vecx_{k+1} = \mathbf{f}_{RK4}\left(\vecx_{k},\vecu_k,dt\right).
\end{equation}
The key challenge here is the numerical integration and discretization techniques over $SO(3)$ for eq.~(\ref{eq:quatdot}). Techniques like the Runge-Kutta method do not necessarily adhere to the $SO(3)$ structure~\cite{Rucker2018}. As a result, the quaternion's norm can become non-unit because of round off error from numerical integration at each time step, which will also induce non-orthogonal column vectors in the corresponding rotation matrix. Extra efforts can be made to circumvent numerical drift by applying a unit constraint. It will add extra constraints to the optimization problem causing additional computational complexity. Another way is to normalize the quaternion after each integration. However, this cannot be done over a horizon in MPC because the result of any intermediate step within the Runge-Kutta calculation is not guaranteed that the unit quaternion represents a $SO(3)$ element. To bypass these strategies, we employ a non-unit quaternion in eq.~(\ref{eq:quatdot}) and its mapping to a rotation matrix is
\begin{equation}~\label{eq:quat}
    \robotrot{} =  
    \frac{\matQ}{\twonorm{\quat{}}^2},
\end{equation}
where $\matQ$ is defined in~\cite{Rucker2018}.
%\begin{equation*}
%   \matQ =  
%    \resizebox{0.9\hsize}{!}{$\begin{bmatrix}
%     q_w^2+q_x^2-q_y^2-q_z^2 & 2\prths{q_xq_y-q_wq_z} & 2\prths{q_wq_y+q_xq_z} \\
%     2\prths{q_wq_z+q_xq_y} &
%     q_w^2-q_x^2+q_y^2-q_z^2 & 2\prths{q_yq_z-q_wq_x} \\
%     2\prths{q_xq_z -q_wq_y} & 2\prths{q_wq_x+q_yq_z} & 
%     q_w^2 - q_x^2-q_y^2+q_z^2\\
%   \end{bmatrix}$}.
%\end{equation*}
The only remaining requirement is to avoid the quaternion getting too close to the zero norm in eq.~(\ref{eq:quat}). It can be shown that eq.~(\ref{eq:quatdot}), which  maps angular velocity to the derivative of a unit quaternion, is a particular solution of the minimum-norm solution for the derivative of a non-unit quaternion~\cite{Rucker2018}. Therefore, to avoid the quaternion magnitude getting close to $0$, we can add to eq.~(\ref{eq:quatdot}) any linear combination of vectors in the nullspace of the weighting matrix of the aforementioned minimum norm problem which turns out to be $\quat{}$ and tune the coefficients. Alternatively, a one time re-scaling step can also be applied.
%\subsection{Practical Implementation}\label{sec:LMPCrelax}

\subsection{LMPC Relaxation}
\subsubsection{Convex Safety Set}\label{sec:convex-ss}
Inspsired by \cite{ROSOLIA20173142}, we approximate the safety set $SS^{j}$ by taking its convex hull $CS^{j}$ as
\begin{equation}
    \begin{aligned}
        CS^j &= Conv(SS^j)=\begin{bmatrix}\vecx{}^0,\cdots,\vecx{}^j\end{bmatrix}\bm{\lambda}^\top, \label{eq:SSbarycentric}
    \end{aligned}
\end{equation}
where 
\begin{align}
    \begin{split}
       &\bm{\lambda} = \begin{bmatrix}
         \lambda^0_0,\lambda^0_1,\cdots,\lambda^0_{T^0},...,\lambda^j_0,\lambda^j_1,\cdots,\lambda^j_{T^j}
       \end{bmatrix},\\
       &\bm{\lambda} \geq 0,\hspace{0.5em} \norm{\bm{\lambda}}_1 =1.\label{eq:sumlambdas}
    \end{split}
\end{align}
Eq.~(\ref{eq:SSbarycentric}) represents the barycentric approximation of $SS^j$. In this way, any state in the convex hull can be written as a convex combination of points in the safety set. Each element in $\bm{\lambda}$ corresponds to a positive weighted scalar value for each state in the convex hull. Similarly, an approximation of the terminal cost function can be derived.
Therefore, we obtain  
\begin{align}
    \begin{split}
        \tilde{Q}^j(\vecx{}) = Conv(Q^j(\vecx{}))&=\min_{\bm{\lambda}\geq0}\left[J^j_{0\xrightarrow{}T^0}(\vecx{}_0^0),J^j_{1\xrightarrow{}T^0}(\vecx{}_{1}^0)\right.,\\
        &~~~~~~~~~~\left. ...,J^j_{0\xrightarrow{}T^j}(\vecx{}_{0}^j),...\right]\bm{\lambda}^\top.
    \end{split}
\end{align}
Using the convex hull of the safety set transforms the NMIP into a Quadratic Program with linear constraints defined by eq.~(\ref{eq:SSbarycentric}) and eq.~(\ref{eq:sumlambdas}).

It should be noted that for sake of simplicity,  in eq.~(\ref{eq:SSbarycentric}) we used the same notation to represent the weighted average operations for both the state variables in the Euclidean space and the variables in the rotation group $SO(3)$. However, in fact, for $SO(3)$ elements, it cannot be achieved in the same way as for elements in the Euclidean space. The notion of Karcher mean~\cite{Hartley2012RotationA} choosing $L2$-norm as metric among two rotation elements guarantees to find the convex set. However, the procedure does not have a closed-form solution and does not guarantee the existence of a unique mean~\cite{Hartley2012RotationA}, thus making it difficult to incorporate it in the MPC. However in our task, the rotations are distributed around the identity element of $SO(3)$ group and we can simply compute the convex safety set $CS^j_{\mathbf{x}_q}$ for the rotation part by lifting all the sample in the tangent space since they lie in the same parameter subgroup~\cite{Moakher2002}  as
$$
CS^j_{\mathbf{x}_q} = \exp\left(\sum\limits_{l=0}^P \lambda_{l}  \log\left(\mathbf{x}_{q,l}\right)\right)
$$
where the $\exp$ and $\log$ operation is defined in~\cite{Sol2017QuaternionKF}, $\mathbf{x}_{q}$ refers to the quaternion vector as a subset of its corresponding state vector, and $l$ indicates a selected state in $CS^j$ among $P$ number of states.
\begin{figure}
    \centering
    \includegraphics[width=1\columnwidth]{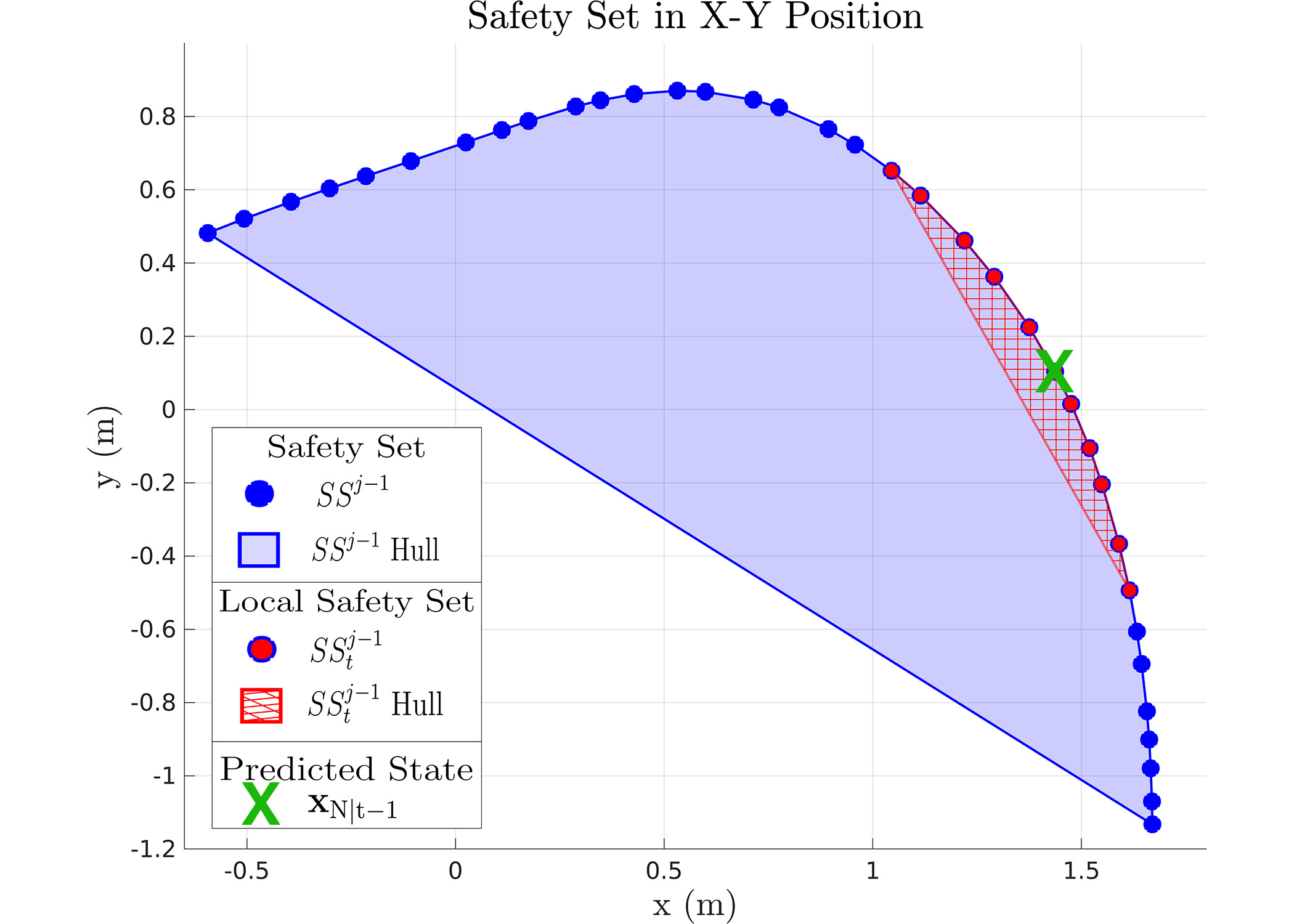}
    \caption{Illustrative example on creating a local safety set in the X-Y position only. At time $t-1$ in $j^{th}$ iteration, the LMPC forecasts states over the horizon, $N$, including the terminal predicted state $\vecx_{N|t-1}$ (green $\times$ in this figure). The local safety set, $SS_t^{j-1}$, at time $t$ (red circles with blue outline) is made up of $\vecx_{N|t-1}$'s nearest neighbors in $SS^{j-1}$. 
     }
    \label{fig:SS_visual}
    \vspace{-20pt}
\end{figure}
\subsubsection{Local Safety Set}\label{sec:local-ss}
In order to further reduce the computational load, we chose to reduce the size of the safety set by choosing a subset, $SS^j_t$, of $SS^j$ at each time $t$
%\begin{align}
%    SS^j_t = \begin{Bmatrix}
%        \vecx{}^i_k ~|~ k \in K^i, i=p,p+1,...,j
%   \end{Bmatrix},
%\end{align}
%\begin{equation*}
%    K^i = \begin{Bmatrix}
%        k^i_1,k^i_2,...,k^i_n
%    \end{Bmatrix}.
%\end{equation*}
%or
\begin{equation}
    SS^j_t =  \left\{ \bigcup\limits^j_{i=p}~\bigcup\limits_{k\in K^i}\vecx{}^i_k\right\}. \label{eq:SStime}
\end{equation}
$K^i = \{k^{i}_1,\cdots,k^{i}_n\}$ is a set of time stamps for the corresponding states in the $i^{th}$ iteration, where $i\in \{p,...,j\}$ and $p\in[0,j-1]$ is an integer that determines how many previous iterations to consider in the safety set. All the sets $K^i$ have an equal size of $n$. The time index in $K^i$ corresponds to the $n$-nearest neighbor states around the predicted terminal state from the predicted trajectory at $t-1$. Fig.~(\ref{fig:SS_visual}) illustrates how the $n$-nearest neighbors are selected for the local safety set. Furthermore, using eq.~(\ref{eq:SSbarycentric}), we approximate the subset by taking the convex hull
\begin{align}
    \begin{split}
        CS^j_t &= Conv(SS^j_t)=\begin{bmatrix}
        \vecx{}^{p}_{k^{p}_1},...,\vecx{}^{p}_{k^{p}_n}...,\vecx{}^{j}_{k^{j}_1} ,...\vecx{}^{j}_{k^{j}_n}
        \end{bmatrix}\bm{\lambda}_s^\top\label{eq:SSlocal}=\vecx{}.
    \end{split}
\end{align}
where
\begin{align}
    \begin{split}
        \norm{\bm{\lambda}_s}_1 =1\label{eq:sumlocallambdas},~
        \bm{\lambda}_s = \begin{bmatrix} \lambda^{p}_{k^{p}_1},...,\lambda^{j}_{k^{j}_n}
        \end{bmatrix}.
    \end{split}
\end{align}
Eqs.~(\ref{eq:SStime})-(\ref{eq:sumlocallambdas}) create a sparse convex hull with non-negative $\bm{\lambda}_s$ elements. 

%\subsubsection{Actuator Constraints}

\subsubsection{LMPC Problem Formulation}
Finally, the relaxed LMPC for quadrotor system can be fully defined as following
\begin{subequations}
\label{eq:LMPCRelax}
    \begin{align}
        \begin{split}
            \tilde{J}^{\scalebox{0.5}{$LMPC$},j}_{t\rightarrow{}t+N}(\vecx_t^j)=\min_{\bm{\lambda}_s,\vecu{}_{t|t},\cdots,\vecu{}_{t+N-1|t}}   \sum_{k=t}^{t+N-1}\hspace{-5pt} h(&\vecx{}_{k|t},\vecu{}_{k|t})\\ &+~\tilde{Q}^{j-1}_{t}(\vecx{}_{t+N|t}),\\ 
        \end{split}
    \end{align}
    \begin{align}
    \text{s.t.}~
        &\vecx_{k+1|t} = \mathbf{f}_{RK4}(\vecx{}_{k|t},\vecu{}_{k|t}),~\forall k \in [t, t+N-1],\\
        &\mathbf{g}(\vecx{}_{k|t},\vecu{}_{k|t})\leq0, ~\vecx{}_{t|t} = \vecx{}^j_t\label{eq:SSrelax_linearconstraint}\\
        &\vecx{}_{t+N|t}\in CS^{j-1}_t,~\norm{\bm{\lambda}_s}_1=1, ~\bm{\lambda}_s\geq0.\label{eq:SSrelaxconstraint}
    \end{align}
\end{subequations}
where
\begin{multline}
    \tilde{Q}^{j-1}_{t}(\vecx{}_{t+N|t})=
    \min_{\bm{\lambda}_s\geq0}\begin{matrix}
        [J^{p}_{k^{p}_1\xrightarrow{}T^{p}}(\vecx{}^{p}_{k^{p}_1}),&J^{p}_{{k^{p}_2}\xrightarrow{}T^{p}}(\vecx{}^{p}_{k^{p}_2}),...\end{matrix}\\
        \begin{matrix}
        J^j_{{k^{j}_1}\xrightarrow{}T^j}(\vecx{}^{j}_{k^{j}_1}),...]
    \end{matrix}\bm{\lambda}_s^\top.
\end{multline}
Our system considers hardware limitations by constraining the control inputs
\begin{align}
    f_{min}\leq & f\leq f_{max}\label{eq:force_constraint},\\
    \bm{\Omega}_{min}\leq&\bm{\Omega}\leq\bm{\Omega}_{max},\label{eq:angvel_constraint}
\end{align}
where $f_{min}$ and $f_{max}$ are the maximum and minimum thrusts, whereas $\bm{\Omega}_{min}$ and $\bm{\Omega}_{max}$ are the maximum and minimum angular velocity, respectively.
\section{Learning Optimal Time Control}~\label{sec:optimaltime}
In this section, we show a particular instance of the LMPC approach to learning optimal time trajectories. This approach can be leveraged for autonomous racing tasks by naturally discovering the minimum lap time through reference-free iterations. Each task begins with a quadrotor at $\vecx{}_s$ which maneuvers to a predefined goal $\vecx{}_G$. A track is created by setting intermediate waypoints and corridors to the goal. An initial safety set, $SS^0$, is created by flying steadily and suboptimally through the track. The LMPC formulation in eq.~(\ref{eq:LMPCform}) is relaxed as discussed in Section~\ref{sec:lmpcquadrotors} and the following cost function

\begin{align}
        \sum^{t+N-1}_{k=t}\left[\mathds{1}\left(\vecx{}_{k|t}\right) + \vecu{}_{k|t}^\top \mathbf{R}_u\vecu{}_{k|t}
        \right]+\tilde{Q}^{j-1}_{t}\left(\vecx{}_{t+N|t}\right) \label{eq:timecost},
\end{align}
where 
\begin{align}
    \mathds{1}\left(\vecx{}\right) = \begin{Bmatrix}
    1, ~&\text{if}~\robotpos{Q} \neq \vecx{}_G \\
    0, ~&\text{Else}
    \end{Bmatrix}.
\end{align}
The running cost is defined by two main components. First, it includes a binary cost which depends on whether the quadrotor has reached the goal position. This cost represents the minimum time control. Second, there is a penalty applied on the inputs to minimize the control effort with $\mathbf{R}_u$ as constant diagonal matrix to tune the penalty on the control. It is important to note that lower values in $\mathbf{R}_u$ favor a minimum-time solution, but this may yield aggressive control and with possible non-smooth control inputs. Therefore, there is a trade-off between these two objectives and tuning the weights is useful to achieving desired performances. While the binary cost can be implemented in simulation, it proves time consuming in real-time application. Thus, eq.~(\ref{eq:timecost}) is approximated with a sigmoid function
\begin{equation}
    h(\vecx{}_{k|t},\vecu{}_{k|t})=\frac{\twonorm{\robotpos{Q,k|t}-\vecx{}_G}^2}{\sqrt{\twonorm{\robotpos{Q,k|t}-\vecx{}_G}^4+1}} + \vecu{}_{k|t}^\top \mathbf{R}_u\vecu{}_{k|t}.
\end{equation}
Furthermore, a track must be defined for the racing case. This can be done with corridors that are defined between two waypoints.
A linear constraint is added to the LMPC formulation to guarantee the quadrotor stays within the track. The position of two distinct consecutive waypoints is defined as $\mathbf{r}_w,\mathbf{r}_{w+1}$, respectively. The direction from the first waypoint to the quadrotor is defined as $\mathbf{r}_{i}=\robotpos{Q}-\mathbf{r}_w$. The normalized direction between each waypoint is $\hat{\mathbf{r}}_c= \frac{\mathbf{r}_{w+1}-\mathbf{r}_w}{\twonorm{\mathbf{r}_{w+1}-\mathbf{r}_w}}$. $\mathbf{r}_n$ is the difference between $\mathbf{r}_i$ and the projection of $\mathbf{r}_i$ onto $\hat{\mathbf{r}}_c$ as
\begin{equation}
\mathbf{r}_n = \left[\textbf{I} - \hat{\mathbf{r}}_c\hat{\mathbf{r}}_c^\top\right]\left(\robotpos{Q} - \mathbf{r}_w\right).
\end{equation}
Physically, $\mathbf{r}_n$ represents the quadrotor's distance to the center axis of the corridor in space. Therefore, a single corridor constraint applies bounds on $\mathbf{r}_n$
\begin{align}
        -\bm{\delta} \leq~ &\mathbf{r}_n \leq \bm{\delta}.
 \label{eq:corridor_constraint}
\end{align}
Therefore, eq.~(\ref{eq:corridor_constraint}) is a linear constraint which matches the form of eq.~(\ref{eq:SSrelax_linearconstraint}) as
\begin{equation}
            \textbf{b}_{min} \leq \textbf{A} \robotpos{Q} \leq \textbf{b}_{max}.
\end{equation}
\section{Experiments}~\label{sec:experimental_results}
We propose several simulations and experiments with a quadrotor to validate the proposed LMPC during a learning minimum time trajectory task.

\subsection{Task Overview}
We consider an optimal time control problem as specified in Section~\ref{sec:optimaltime}. The task is successfully accomplished if the quadrotor can pass through a given gate while staying within a given track and stop after passing it. Fig.~\ref{fig:intro} illustrates an example of the proposed task. This is obtained considering an L-shape track, which is constructed using the approach in Section.~\ref{sec:optimaltime} both in simulations and real-world experiments.
We first provide the quadrotor with a feasible and slow reference trajectory for the quadrotor MPC controller to track, like the one shown in the black color in the Figs.~\ref{fig:sim-traj-hist} and \ref{fig:real-world-traj-hist}. During the trajectory tracking, the quadrotor records its state history to build the initial safety set. Subsequently, we run the LMPC controller and repeat the same process by sending the quadrotor to the same start position at the end of each task iteration. The only information provided to LMPC is the safety set and the corresponding cost-to-go over the recorded safety set. The LMPC will then start to find the optimal time trajectories that minimize the travel time while respecting the system dynamics, actuator constraints, and track constraints. The attached multimedia material provides several additional experiments as well. We solve the proposed LMPC problem in eq.~(\ref{eq:LMPCRelax}) with cost as eq.~(\ref{eq:timecost}) via Sequential Quadratic Programming (SQP) using ACADOS~\cite{Verschueren2018,Verschueren2019} as a solver. For the LMPC controller parameters, we choose the prediction time horizon the discretized time step $dt$ as $0.1~\si{s}$, the horizon length $N = 10$ and corridor width $\bm{\delta} = 0.8~\si{m}$.

\subsection{Environments}
\subsubsection{Simulation} For simulation, we use a custom simulator available in the lab developed in ROS\footnote{\url{www.ros.org}} with full system dynamics simulated using $4^\text{th}$ order Runge-Kutta method. 
\subsubsection{Real World}
The real-world experiments are performed in an indoor testbed with a flying space of $10\times6\times4~\si{m^3}$ at the ARPL lab at the New York University. We leverage a Vicon\footnote{\url{www.vicon.com}} motion capture system at $100~\si{Hz}$ for control purposes. The quadrotor platform setup is similar to our previous work~\cite{LoiannoRAL2017}. The control and estimation frameworks are developed in ROS. The proposed LMPC method can run on-board at $100~\si{Hz}$ on a common laptop. 
\subsection{Results}
\begin{figure}[t]
    \centering
    \includegraphics[width=\columnwidth]{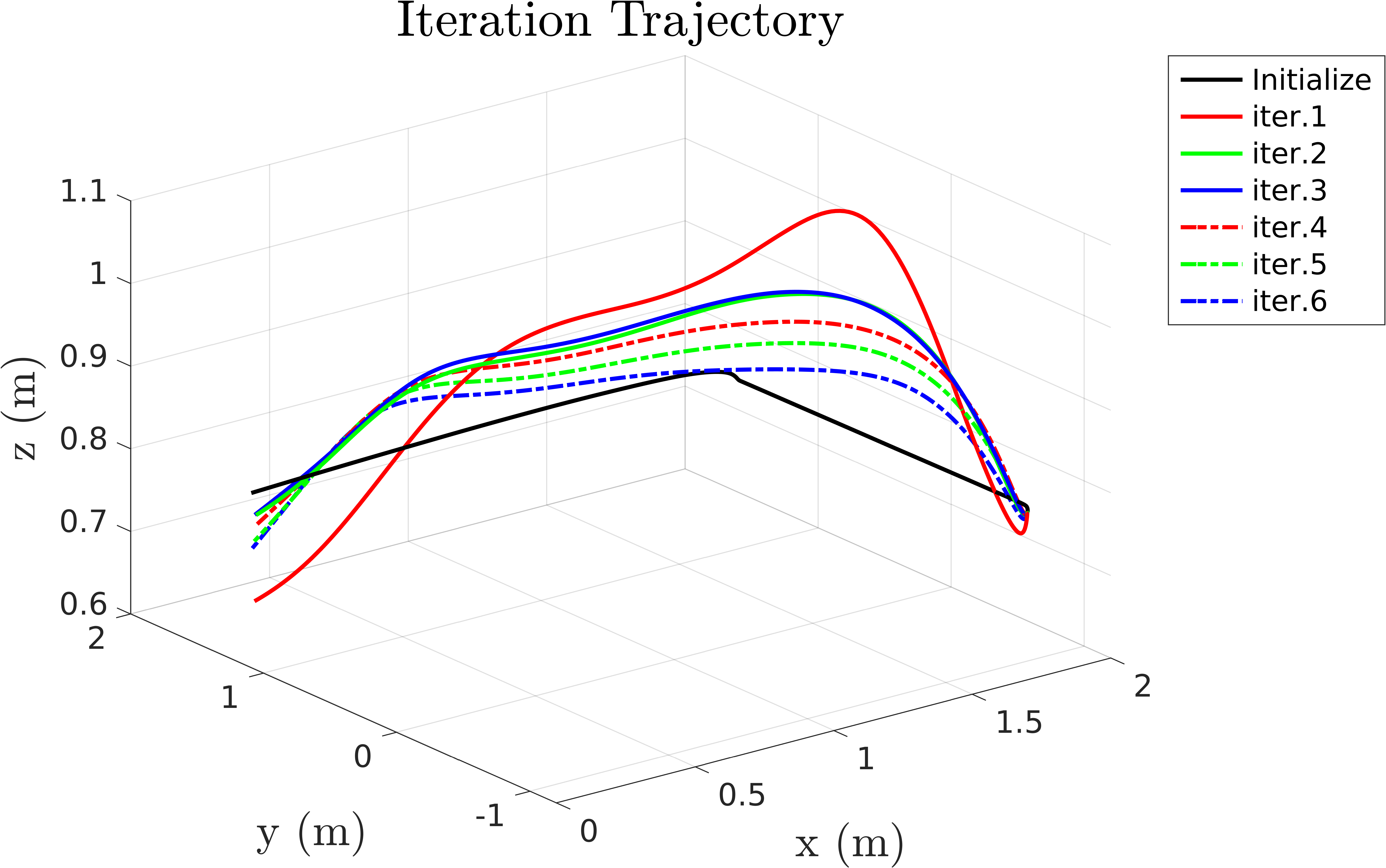}
    \caption{Iteration trajectories for the L-shaped track in simulation.}
    \label{fig:sim-traj-hist}
        \vspace{-10pt}
\end{figure}

\begin{figure}[t]
    \centering
    \includegraphics[width=\columnwidth]{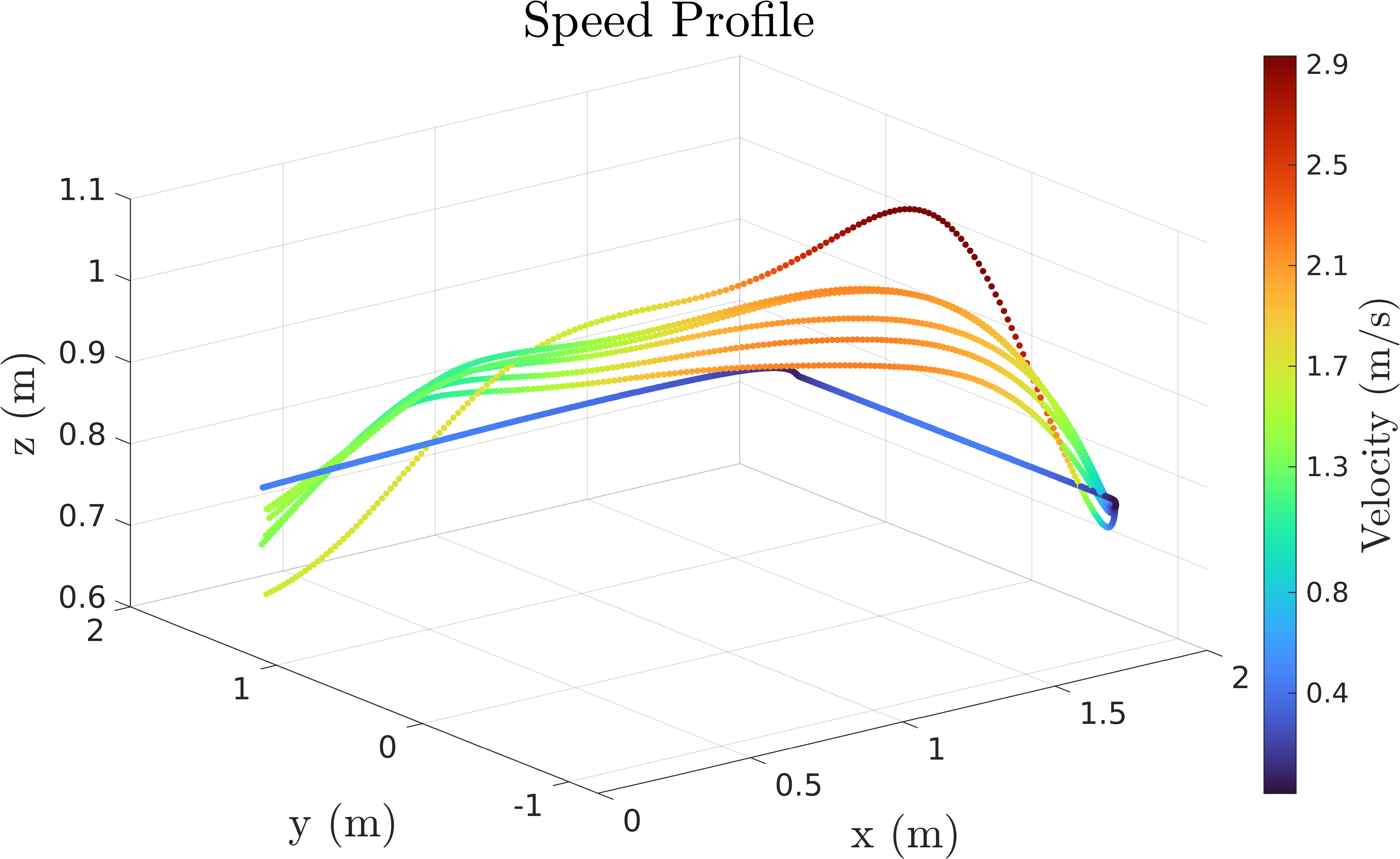}
    \caption{Trajectory and velocity profiles across several iterations for the L-shaped track in simulation.}
    \label{fig:sim-speed-profile}
    \vspace{-10pt}
\end{figure}

\begin{figure}[!t]
    \centering
    \includegraphics[width=\columnwidth]{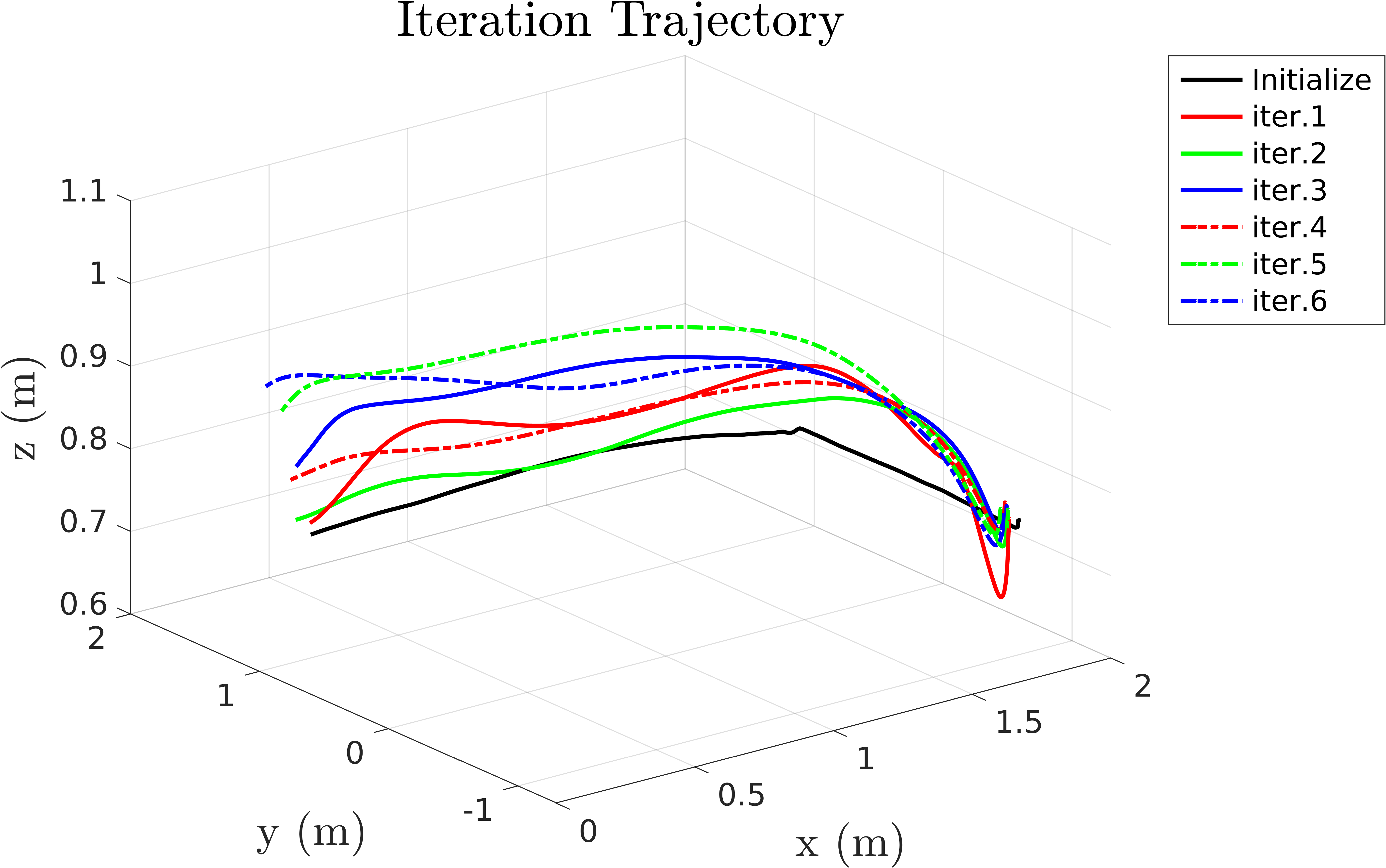}
    \caption{Iteration trajectories for the L-shaped track in real-world experiments.}
    \label{fig:real-world-traj-hist}
    \vspace{-10pt}
\end{figure}

\begin{figure}[h]
    \centering
    \includegraphics[width=\columnwidth]{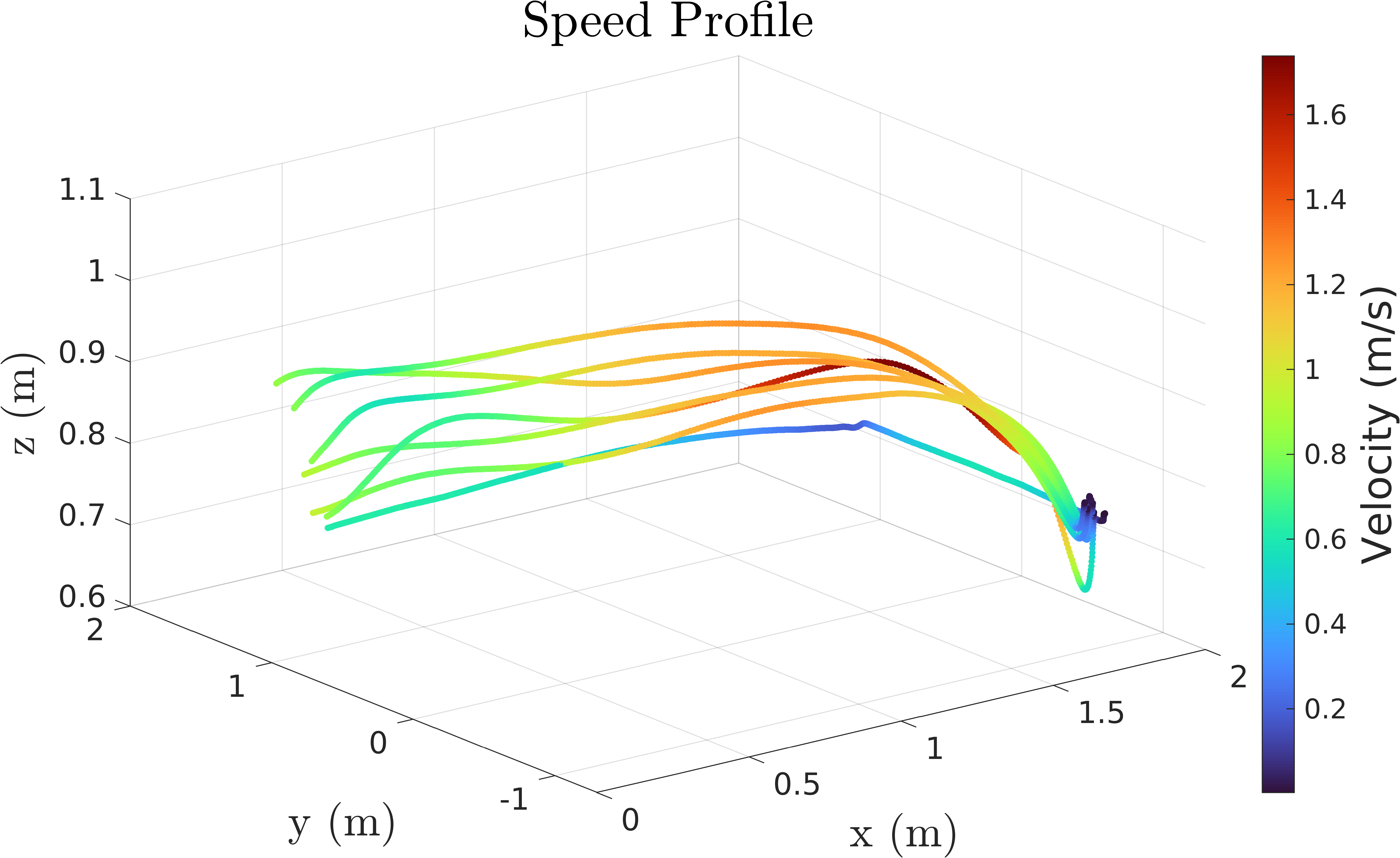}
    \caption{Trajectory and velocity profiles across several iterations for the L-shaped track in real-world experiment.}
    \label{fig:real-world-speed-profile}
       \vspace{-10pt}
\end{figure}

We show the simulation and real-world test results of a quadrotor traveling through an L-shape track with the LMPC.
The results of the simulation are shown in Figs.~\ref{fig:sim-traj-hist} and \ref{fig:sim-speed-profile} whereas the real-world experiments in Figs.~\ref{fig:real-world-traj-hist} and \ref{fig:real-world-speed-profile}. The travel time for each iteration is reported in Table~\ref{tab:travel_time}. As we can observe from the plots, both in the simulation and real-world experiments, the quadrotor can explore the track and find a locally optimal time trajectory which ends up a much faster trajectory than the initial one.
In Table~\ref{tab:travel_time}, we observe that as the iterations continue, the LMPC converges similar and stable travel time along the track. This proves that the LMPC can utilize the recorded states and costs in the past iteration and explore new faster trajectories for the quadrotor to accomplish the task. However, we notice that the final lap time varies around a given value after convergence. We believe this behavior is mostly due to a mismatch between real and modeled dynamics including our approximation to first-order attitude dynamics in eq.~(\ref{eq:quatdot}).

%It is important to note that lap time does not strictly decrease with successive iterations in both real and simulation experiments. This is most likely a result of a mismatch or limitation of the model dynamics. Our system considers up to the first order attitude dynamics which may limit the desired behavior. A more comprehensive approach may use attitude dynamics up to the second order. However, this inherently increases the computational complexity which may affect the performance in a real-time setting. This trade off will be explored in the future.

% \begin{figure}[!t]
%     \centering
%     \includegraphics[width=\columnwidth]{figs/experiments/SimIter_ICRA_2022.png}
%     \caption{Iteration trajectories for the L-shaped track in simulation.}
%     \label{fig:sim-traj-hist}
%         \vspace{-10pt}
% \end{figure}
% \begin{figure}[h]
%     \centering
%     \includegraphics[width=\columnwidth]{figs/experiments/SimIter_SpeedProfile_ICRA_2022.png}
%     \caption{Trajectory and velocity profiles across several iterations for the L-shaped track in simulation.}
%     \label{fig:sim-speed-profile}
%     %\vspace{-10pt}
% \end{figure}

\begin{table}[!t]
%\vspace{-10pt}
\caption{The total travel time (s) of different iterations in simulation and real-world experiments \label{tab:travel_time}} 
\centering
%\newcolumntype{s}{}
\begin{tabularx}{\columnwidth}{>{\hsize=0.25\hsize}X >{\hsize=0.25\hsize}X >{\hsize=0.25\hsize}X >{\hsize=0.25\hsize}X >{\hsize=0.25\hsize}X >{\hsize=0.25\hsize}X >{\hsize=0.25\hsize}X >{\hsize=0.25\hsize}X}
    \hline\hline
 \rule{0pt}{2ex} 
 Iter & Init  & 1 & 2 & 3 & 4 & 5 & 6 \\\hline
 Real & 10.0002 &   3.5291  &  3.6993  &  3.9895   & 3.6296  &  3.7494  &  3.7292 \\  
  Sim & 14.1291  &  2.0900   & 2.4000  &  2.4300  &  2.4600  &  2.5299  &  2.5800\\
  \hline\hline
\end{tabularx}
\vspace{-10pt}
\end{table}

\section{Conclusion}~\label{sec:conclusion}
In this paper, we presented an LMPC for quadrotors. We addressed the challenges associated with the system evolution on a nonlinear manifold configuration space which requires careful considerations in the LMPC problem formulation as well as in its forward numerical time integration. We showed how to reduce its computational complexity to make it compatible with the stringent requirements for real-time control of quadrotors as well as its usefulness in a learning minimum time trajectory problem.

Future works will investigate the trade-offs between the incorporation of the dynamics till rotor speeds in the proposed approach to improve the system's performance, agility, and the corresponding increase in computational complexity which may affect the system real-time performances. We will leverage Bayesian machine learning techniques to refine the system dynamics incorporating unmodelled dynamical effects across multiple runs thus allowing us to further push the system performances and agility limits.
We will also consider employing this method for drone racing tasks extending the proposed experiments to a full racing track. Finally, we will investigate the use of different cost functions and how the availability of reference trajectories can potentially be exploited to improve the task performances.

% \todo{adjust this param here}
\addtolength{\textheight}{-10.0cm}   % This command serves to balance the column lengths

%\section*{APPENDIX}

\bibliographystyle{IEEEtran}	% (uses file "plain.bst")
\bibliography{references}
%\biboptions{sort&compress}
%\input{10-old.tex}
\end{document}